%% file: root.tex
\documentclass{isprs} 
\usepackage{setspace}
\usepackage{geometry} 
\usepackage{epstopdf}
\usepackage[labelsep=period]{caption}  
\usepackage[british]{babel} 
\usepackage[hang]{footmisc}
\usepackage{amsmath}
\usepackage{url}
\usepackage{amssymb}
\usepackage[hidelinks]{hyperref}
\usepackage{csquotes}
\usepackage[authoryear]{natbib}

\usepackage[caption=false,font=footnotesize]{subfig}
\usepackage[nolist]{acronym}
\begin{document}
\input{acronym} 

\title{Supercharging Thermal Gaussian Splatting with Depth Estimation}
\date{}

\author{
 Manoj Biswanath\textsuperscript{1}\thanks{These authors contributed equally to this work.}  , Chenxin Cai\textsuperscript{2}\footnotemark[1]  , Hannah Schieber\textsuperscript{3}, Daniel Roth\textsuperscript{3}, Benjamin Busam\textsuperscript{1} }

\address{
	\textsuperscript{1 }Photogrammetry and Remote Sensing, Munich Center for Machine Learning (MCML),\\ Technical University of Munich, Munich, Germany - (manoj.biswanath, b.busam)@tum.de\\
	\textsuperscript{2 }Technical University of Munich, Munich, Germany - chenxin.cai@tum.de\\
	\textsuperscript{3 }Human-Centered Computing and Extended Reality Lab, TUM University Hospital,\\Clinic for Orthopedics and Sports Orthopedics, Munich Institute of Robotics and Machine Intelligence (MIRMI),\\Technical University of Munich, Munich, Germany – (hannah.schieber, daniel.roth)@tum.de\\
}



\abstract{
Efficient and robust 3D scene representation is crucial in autonomous driving, robotics, and related fields. While RGB images provide valuable content for 3D reconstruction,
other modalities like thermal or depth can enable additional
information on the environment. Lately, novel view synthesis methods like 3D Gaussian Splatting have started using multiple modalities to further boost their performance. But fusing or combining multimodal data can
make the process slower and can bring in additional challenges.
Therefore, our project aims to use single modality based on thermal infrared domain, by removing the reliance on visible
light as much as possible. This single
modality can be expected to be faster as it does not rely
on multimodal data. We propose a method, \textbf{\ac{tdg}}, that uses only thermal images and
depth estimation in its architecture to derive the radiance fields. Our \ac{tdg} method outperforms the MSMG (Multiple Single-Modal Gaussians) baseline in most cases on our test datasets, RGBT-Scenes and ThermalMix. On average,
the rendering quality metrics such as \ac{lpips}, \ac{ssim}, and \ac{psnr} of \ac{tdg} are 1.12\%, 0.034\%, and 0.01\% better than the baseline MSMG values. It also reduces the training time significantly, by 12 mins 47 secs (55\% improvement). Overall, our method is successful
in deriving these thermal radiance fields, which can
ultimately have several applications, such as identifying heat
sources critical in surveillance, search or rescue operations, and industrial inspections where temperature
is widely used to monitor machines.  
}


\maketitle


\section{Introduction}

Robotics \citep{ji_graspsplats_2024,li_object-aware_2024}, autonomous driving \citep{yan2024street}, virtual reality \citep{schieber_semantics-controlled_2024,kleinbeck2026hybridfoveatedpathtracing}, construction industry \citep{ruter20243d} and similar fields demand highly efficient and robust 3D scene representation.  In recent years, explicit rendering techniques, particularly \ac{nvs} methods such as \ac{gs} \citep{kerbl_3d_2023}, received significant attention due to their exceptional rendering speed and visual quality \citep{li2025radiance}. Mainstream methods heavily rely on RGB images, whose performance deteriorates significantly in visually degraded environments, such as low-light, fog, smoke, or extreme weather conditions. This can make the direct application of RGB-designed \ac{gs} frameworks highly challenging. Introducing other modalities alongside RGB, such as thermal, depth, or LiDAR data, can provide additional information about the environment, or can especially help in cases where drones and ground robots are often equipped with additional sensors, or only contain thermal and depth cameras, rather than RGB.

\begin{figure}[t!]
    \centering
    \includegraphics[width=1.006\linewidth]{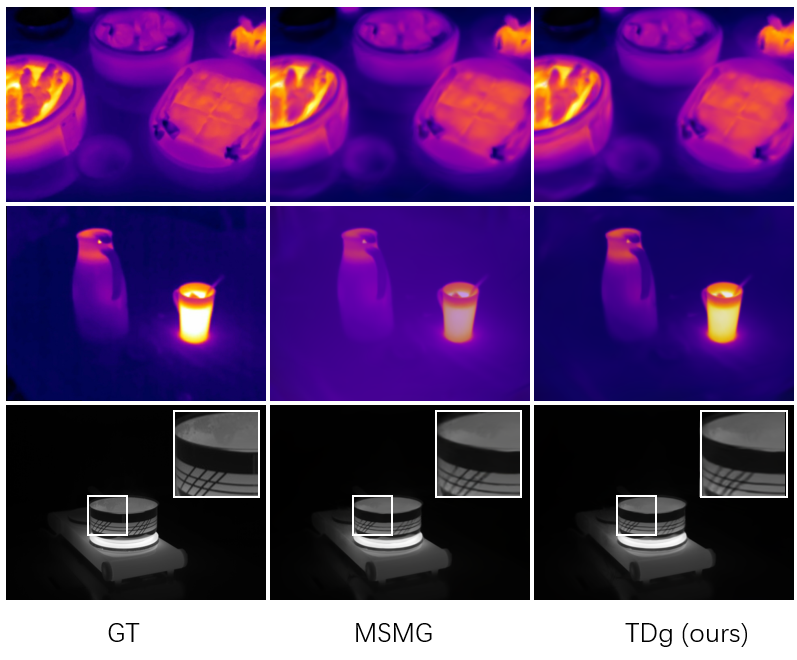}
\caption{%
Thermal images rendered from the \ac{gs} model of our \ac{tdg} method, compared against Ground Truth (GT) and baseline (MSMG).%
\label{fig:comparison}%
}
\end{figure}

Infrared imaging cameras, which detect the inherent thermal radiation of objects, can provide a robust perception capability, independent of visible light. It is suitable in challenging environments, regardless of lighting and weather conditions. Thermal imaging converts temperature information into interpretable images. One common application of thermography is energy inspection of objects. Classically, this has been done by visually evaluating the images. But a direct 3D spatial reference is not established for the measured values. This deficiency becomes obvious for complex structures, when their images taken from different angles need to be fused, or the measured thermal radiations need to be stored in an object-related manner, for further processing. In such cases, creation of a thermal radiance field (as a 3D representation) can be helpful. It can provide us a 3D spatial reference of the temperature measurements. For example, one specific application could be energy inspection in buildings, where knowledge of both the inside and outside wall temperatures would be necessary to provide insights about the geometry and material properties of the walls. This knowledge could be acquired from the 3D representations. Further, these continuous scene representations can provide correct view-dependent effects such as reflections, translucency, etc., and can serve as either digital twins or an enrichment of already available digital twins.

Novel views are required to get the correct heat signatures of objects, avoiding parallax errors. Thermal patterns of objects can be very complex. For example, heat leakages from building walls mostly look like blobs, elongated structures, or irregular shapes \citep{krawczyk2015infrared}. To obtain a more accurate picture of these complex patterns, many different viewpoints would be necessary to have the ability to rotate, view, and inspect them. Novel views can provide realistic images from viewpoints that were never actually captured, because taking photos from all angles is not feasible. Also, continuous viewpoint changes can be useful to study space/time-related heat transfer behavior. And predict heat radiation from new viewpoints.

Recent studies have attempted to integrate thermal imaging into the GS framework by using multiple modalities. This can further boost their performance by providing accurate information, enabling an exploration of the 3D environment, and providing photorealistic results. But fusing or combining these multi-modal data can make the process slower and can bring in additional challenges. Leveraging a single modality eliminates this dependency, enables faster convergence, and simplifies the processing pipeline. Thermal and RGB cameras typically require precise co-registration or alignment. Thermal images are inherently characterized by low contrast, sparse texture, and non-uniform brightness distribution. Current approaches still fundamentally rely on paired RGB images for supervision or joint optimization, failing to establish a truly independent and purely thermal-based Gaussian representation system. Since RGB images can enhance the quality of thermal models because of their rich geometry and semantic information, removing the reliance on RGB while still maintaining good quality is challenging and demands an intelligent architecture.

Therefore, to address the challenges of thermal-based \ac{gs} fundamentally, we propose an enhanced and self-contained Thermal \ac{gs} framework. We want to build a \ac{gs} model to synthesize thermal images. Through core innovation, our framework achieves high-quality 3D reconstruction. In summary, we contribute:

\begin{enumerate}
\item A thermal-guided depth estimation module integrated within \ac{gs}
\item A novel purely thermal \ac{gs} approach, denoted \textbf{\acf{tdg}}\footnote{Code available: \href{https://hannahhaensen.github.io/TDg/}{https://hannahhaensen.github.io/TDg/}}
\item A strong evaluation on two state-of-the-art benchmark datasets, RGBT-Scenes and ThermalMix.
\end{enumerate}

\begin{figure*}[t!]
    \centering
    \includegraphics[height=0.52\linewidth]{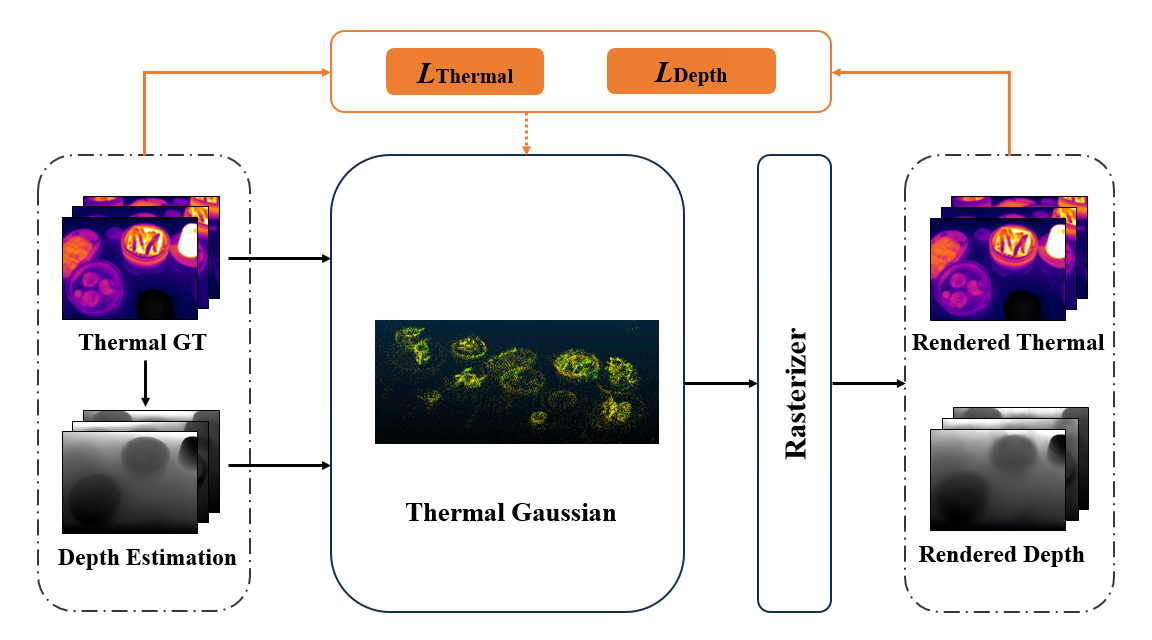}
    \caption{\ac{tdg} architecture. A unified 3D Gaussian representation (center) is optimized via dual rasterization. By comparing the rendered thermal and depth maps (right) against the input thermal GT and estimated depth priors (left), a joint loss (top) is computed to backpropagate and guide the geometric and photometric training of the Gaussians.}
    \label{fig:architecture}
\end{figure*}

Figure~\ref{fig:comparison} shows a teaser of our method's results. The thermal images rendered from our resulting GS model, are compared against the baseline method (MSMG) and Ground Truth (GT). As shown in the comparison, our method shows better resolution. 

\section{Related Work}

Since our approach integrates depth estimation with thermal \ac{nvs}, it lies at the intersection of geometric scene understanding and multi-modal 3D reconstruction. Consequently, we review prior work on depth estimation methods as well as approaches that extend \ac{nvs} to multi-modal or cross-spectral settings. Particular emphasis is placed on techniques that leverage complementary sensing modalities to improve geometric and photometric consistency.

\subsection{Novel View Synthesis}

Early approaches such as \ac{gs} \citep{kerbl_3d_2023} and \ac{nerf} \citep{mildenhall_nerf_2022} rely on RGB images and camera poses typically provided by COLMAP \citep{schonberger_structure--motion_2016}. While later works improve joint camera optimization \citep{schischka_dynamon_2024,schieber_nerftrinsic_2024,bian_nope_2023}, COLMAP-based initialization remains common.
Recent research extends radiance fields to the thermal domain \citep{chen2024thermal3d,lu_thermalgaussian_2024,carmichael2025trnerf,hassan_thermonerf_2025}, enabling 3D reconstruction beyond the visible spectrum. 

The most frequent multi-modal approaches add depth data instead of thermal images. For example, \citep{chung_depth_2024} leverages dense depth maps aligned with initial COLMAP data to optimize for a more precise geometry of the scene. Boosting \ac{gs} performance with depth is further investigated by \citep{Xu_depthsplat_2025}. They combine depth estimation and \ac{gs} in a shared architecture. Both show advantages for \ac{nvs}.

\subsection{Thermal Novel View Synthesis}

Similar to recent developments in general \ac{nvs}, both \ac{gs} \citep{kerbl_3d_2023} and \ac{nerf} \citep{mildenhall_nerf_2022} have also demonstrated strong applicability to thermal \ac{nvs}. Thermal-\ac{nerf} \citep{ye_thermal_2024} leverages infrared imagery to enhance rendering in low-light environments: their method converts 16-bit infrared inputs into 8-bit thermal representations, defines a fixed-size sampling volume, and employs an MLP to regress thermal intensities. While Thermal-\ac{nerf} focuses solely on thermal inputs, Thermo\ac{nerf} \citep{hassan_thermonerf_2025} integrates RGB and thermal modalities through two coupled MLPs. In a similar direction, \citep{ozer_exploring_2024} proposes a joint training framework for RGB and thermal images, systematically exploring different integration strategies and evaluating reconstruction quality across both modalities. \citep{carmichael2025trnerf} introduces TRNeRF, a framework designed to restore blurry, rolling-shutter, and noisy thermal imagery; their approach explicitly considers the various types of thermal cameras and the distinct challenges they introduce, such as the motion blur commonly observed in uncooled microbolometers, addressing these issues by using models of rolling-shutter timing, noise having fixed-patterns, and motion blur directly into NeRF.

\citep{lu_thermalgaussian_2024} extends \ac{gs} to the multimodal \ac{nvs} setting by jointly training on RGB and thermal images simultaneously, rendering Gaussians separately for each modality while optimizing them through a combined loss. In this way, they allow each Gaussian to specialize in rendering its respective modality. They use a shared, multimodal-initialized point cloud and also introduce multimodal regularization based on the number of Gaussians associated with each modality. They introduce three different training strategies. One of them is MSMG (Multiple Single Modal Gaussians), which we select as the baseline method for comparison of our work. The reason for selecting this method as our baseline is because it closely resembles our proposed method, in the sense that our method replaces the RGB modality with purely thermal images advanced with depth estimation. Additionally, this method demonstrates fairly good results in thermal-GS domain, which is ideal for baseline comparison. \citep{nam2025veta} presents Veta-\ac{gs}, a thermal \ac{gs} approach that is aware of scene dynamics. By integrating a Thermal Feature Extractor and a deformation field which is view-dependent, the method captures thermal variations while maintaining robustness, and additionally incorporates the MonoS-SIM loss, which emphasizes frequency, appearance, and edge information to preserve stability.
Beyond RGB–thermal fusion, \citep{chen2024thermal3d} proposes a physics-inspired thermal \ac{gs} framework. Their method explicitly models thermal conduction effects, atmospheric transmission, and incorporates a temperature consistency constraint tailored to infrared signal behavior.

\subsection{Depth Estimation}

A wide range of approaches has been proposed for estimating depth from RGB images \citep{arampatzakis_monocular_2023,ke_repurposing_2024}. Recently, depth estimation via diffusion-based techniques has shown particularly strong performance \citep{ke_repurposing_2024}. They adapt Stable Diffusion to the monocular depth estimation task by fine-tuning its U-Net backbone for depth prediction. \citep{chung_depth_2024} leverages dense depth maps aligned with initial COLMAP data to optimize for a more precise geometry of the scene.

\paragraph{Thermal Depth Estimation} In addition to RGB-based methods, thermal depth estimation provides an alternative line of research. Marigold~\citep{ke_repurposing_2024}  uses thermal images to estimate depth maps. Enhancing thermal images by maximizing self-supervision has been attempted by \citep{Shin2022}. They demonstrate the importance of temporally consistent normalization and adopt contrast
limited adaptive histogram equalization (CLAHE) \citep{Zuiderveld1994Clahe} to intensify image details locally, by abstaining from noise amplification. \citep{shin_deep_2023} and \citep{zuo_monother_2025} investigate depth estimation directly from thermal imagery, demonstrating its potential, especially under difficult lighting conditions or in adverse environments.

\paragraph{Depth Estimation and Novel View Synthesis} Depth estimation can be incorporated into \ac{nvs} pipelines to further enhance reconstruction quality and overall robustness, e.g. \citep{Xu_depthsplat_2025}. For improving \ac{nvs}, depth can be a valuable factor. Traditional \ac{gs} \citep{kerbl_3d_2023} allows integration of depth within further code refinement to supervise \ac{gs}. While often depth-supervised methods focus on the center of Gaussians and ignore the unique Gaussian shape during training, SAD-\ac{gs}~\citep{kung2024sad} directly addresses this problem explicitly, by enhancing \ac{gs} with depth supervision. Another work specifically optimizing depth and \ac{gs} is, for example, introduced in DepthSplat \citep{Xu_depthsplat_2025}. DepthSplat connects \ac{gs} with single and multi-view depth estimation, resulting in superior-quality \ac{gs}. It uses pre-trained monocular depth features. \citep{Roessle_2022_CVPR} use dense depth priors for deriving the neural radiance fields. They reduce the required number of images significantly by an order of magnitude, to synthesize the novel views.

\subsection{Research Gap}
We have seen how research has evolved from initial RGB-based \ac{nvs} to accommodating multiple modalities or cross-spectral domains. Different attempts are seen to integrate thermal into \ac{nvs} using various fusion techniques. Eventually, some thermal-based \ac{gs} and \ac{nerf} are developed. Additionally, people tried to use depth to supervise \ac{nvs}. They explored different depth estimation techniques, both from RGB and thermal modalities. While all these methods tried to integrate multi-modal data, but they rather treated these additional modalities as complementary, mostly for improvement purposes. Eventually, none of them developed a purely single modality (thermal infrared) based \ac{gs} framework. So, we have attempted to bridge this gap through our work. We also use the depth factor for supervision and the depth is estimated solely from thermal images. Therefore, our attempt is to retain our framework within a single modality.

\section{Method}

We introduce our enhanced thermal \ac{gs} framework \ac{tdg}. We replace the RGB modality, and purely rely on thermal images advanced with depth estimation, purely queried by thermal images, see Figure \ref{fig:architecture}. This method significantly improves the geometric accuracy and radiance consistency of the thermal scene representation.

In our \ac{tdg} framework, rather than maintaining separate models for different modalities, we utilize a single, unified set of 3D Gaussians to inherently align the thermal radiance and geometry of the scene. Specifically, a set of $N$ Gaussians $\mathcal{G}$ is used to represent the scene:
\begin{equation}
    \mathcal{G} = \{ \mu_i, \Sigma_i, \alpha_i, c_i \}_{i=1}^{N}
\end{equation}
where each Gaussian $i$ is parameterised by its covariance matrix $\Sigma_i$, a thermal radiance feature $c_i$, center position $\mu_i$, and  opacity $\alpha_i$. During optimization, this unified model is rendered into both thermal images (using the radiance feature $c_i$) and depth maps (using the camera-space $Z$-coordinate of $\mu_i$). This design allows the pre-estimated depth prior to directly constrain the 3D geometric distribution of the thermal radiance field, avoiding the complexity and feature conflicts of multi-branch architectures.

\subsection{Thermal-to-Depth Gaussian Estimation}

Building upon the \ac{gs} rasterization strategy \citep{kerbl_3d_2023}, we remove the reliance on RGB and integrate depth instead. Our approach purely leverages estimated depth from thermal images. This explicitly enforces geometric constraints and addresses the instability of pure thermal supervision in low-texture or low-light conditions.  

\subsubsection{Thermal-to-Depth Estimation}

Given a thermal image $I_{Th}$, a thermal-guided depth estimator $f_\theta$ predicts a dense depth map:
\begin{equation}
    D = f_\theta(I_{Th}).
\end{equation}

\subsubsection{Depth Gaussian Loss}

Similar to the color rendering of thermal images, the rendered depth map $\hat{D}$ is obtained via $\alpha$-blending integration along the ray direction. For a given pixel $i$, its expected depth $\hat{D}_i$ is accumulated based on the $Z$-axis depth $z_k$ of the Gaussians in the camera coordinate system:
\begin{equation}
    \hat{D}_i = \sum_{k \in \mathcal{N}_i} z_k \alpha_k \prod_{j=1}^{k-1} (1 - \alpha_j),
\end{equation}
$\mathcal{N}_i$ represents the arranged set of overlapping Gaussians along the ray corresponding to pixel $i$. 

Since the depth predicted by monocular depth estimation networks typically contains scale and shift ambiguities, it cannot be directly compared with the absolute rendered depth. Therefore, we apply Min-Max normalization to both rendered depth $\hat{D}$ and predicted depth $D$, mapping them into a unified scale $[0, 1]$. Let $\tilde{D}$ and $\tilde{\hat{D}}$ represent normalized estimated and rendered depth maps, respectively. The depth loss is formulated as a combination of \ac{ssim} (Structural Similarity) and $L_1$ distance:
\begin{equation}
    \mathcal{L}_{\text{depth}} = \| \tilde{\hat{D}} - \tilde{D} \|_1 + (1 - \text{\ac{ssim}}(\tilde{\hat{D}}, \tilde{D})).
\end{equation}

\subsubsection{Thermal Gaussian Loss}

Following \citep{lu_thermalgaussian_2024}, the thermal supervision combines pixel fidelity, structural similarity, and a smoothness term penalizing local discontinuities:
\begin{equation}
\begin{split}
    \mathcal{L}_{\text{thermal}} ={} & (1-\lambda_{\ac{ssim}})\mathcal{L}_1(I_{Th}, \hat{I}_{Th}) \\
    & + \lambda_{\ac{ssim}} (1 - \text{\ac{ssim}}(I_{Th}, \hat{I}_{Th})) + \lambda_{smooth}\mathcal{L}_{\text{smooth}},
\end{split}
\end{equation}
where $\mathcal{L}_1$ denotes the mean absolute error between the ground-truth thermal image $I_{Th}$ and the rendered thermal image $\hat{I}_{Th}$. It is used to assess the geometric accuracy.     

$\mathcal{L}_{\text{smooth}}$ is a smoothness loss term introduced for regularization:
\begin{equation}
    \mathcal{L}_{\text{smooth}} = \frac{1}{4M}\sum_{i,j} (|T_{i\pm1,j} - T_{i,j}| + |T_{i,j\pm1} - T_{i,j}|),
\end{equation}
where $M$ is the number of rendering pixels in total and $T_{i,j}$ is the temperature at pixel $(i,j)$. Because objects above absolute zero continuously radiate heat and reach thermal equilibrium with their surroundings, most regions in thermal images exhibit gradual temperature changes. Hence, introducing the smoothness term effectively penalizes unnatural abrupt changes and noise.

\begin{table*}[t!]
    \centering
    \begin{tabular}{|l|cccc|cccc|cccc}
        \hline 
         \rule{0pt}{0ex} & & & & & & & & \\
        \textbf{Datasets}
        & \multicolumn{4}{c|}{\textbf{MSMG \citep{lu_thermalgaussian_2024}}} 
        & \multicolumn{4}{c|}{\textbf{Ours (\ac{tdg})}}
         \\
        & \ac{psnr} $\uparrow$ & \ac{ssim} $\uparrow$ & \ac{lpips} $\downarrow$ & Time [h] $\downarrow$
        & \ac{psnr} $\uparrow$ & \ac{ssim} $\uparrow$ & \ac{lpips} $\downarrow$ & Time [h] $\downarrow$
        \\ 
        \hline 
        \rule{0pt}{0ex} & & & & & & & & \\
         \textbf{RGBT-Scenes}  & & & & & & & &
         \\[3pt]
         \textbf{~~Day light conditions}  & & & & & & & &
         \\
        ~~~~~Dimsum & 26.80 & 0.893 & 0.127 & 0.407 & \textbf{26.90} & \textbf{0.895} & \textbf{0.121} & \textbf{0.376} \\
        ~~~~~DailyStuff & 21.04 & 0.830 & 0.203 & 0.402 & \textbf{21.17} & \textbf{0.836} & \textbf{0.199} & \textbf{0.106} \\
        ~~~~~Ebike & 23.20 & \textbf{0.872} & 0.206 & 0.390 & \textbf{24.99} & 0.865 & 0.237 & \textbf{0.289} \\
        ~~~~~Road Block & 26.51 & 0.921 & 0.221 & 0.388 & \textbf{26.73} & \textbf{0.921} & \textbf{0.212} & \textbf{0.159} \\
        ~~~~~Truck & 25.90 & 0.875 & 0.145 & 0.384 & \textbf{26.28} & \textbf{0.878} & \textbf{0.133} & \textbf{0.227} \\
        ~~~~~Rotary Kiln & {26.80} & 0.928 & 0.145 & 0.391 & \textbf{26.85} & \textbf{0.929} & \textbf{0.134} & \textbf{0.105} \\
        ~~~~~Parterre & 24.06 & 0.877 & 0.219 & 0.385 & \textbf{24.42} & \textbf{0.879} & \textbf{0.205} & \textbf{0.118} \\
        ~~~~~Glass Cup & 25.12 & 0.861 & 0.129 & 0.387 & \textbf{25.66} & \textbf{0.873} & \textbf{0.124} & \textbf{0.102} \\
        [3pt]
        \textbf{~~Dark light conditions}  & & & & & & & &
         \\
        ~~~~~Dark Scenes & 19.40 & 0.811 & 0.322 & 0.359 & \textbf{19.84} & \textbf{0.815} & \textbf{0.307} & \textbf{0.104} \\
        [6pt]
        \textbf{~~Average} & 24.31 & 0.874 & 0.191 & 0.388 & \textbf{24.76} & \textbf{0.877} & \textbf{0.186} & \textbf{0.176} \\
        \hline    
        \rule{0pt}{0ex} & & & & & & & & \\
         \textbf{ThermalMix}  & & & & & & & &
         \\[3pt]
        \textbf{~~360-degree scenes}  & & & & & & & &
         \\
        ~~~~~Laptop & \textbf{30.03} & \textbf{0.912} & \textbf{0.190} & 0.388 & 29.88 & 0.904 & 0.202 & \textbf{0.102} \\
        ~~~~~Pan & 32.15 & \textbf{0.897} & 0.068 & 0.407 & \textbf{32.63} & 0.887 & \textbf{0.066} & \textbf{0.284} \\
        ~~~~~Lion & \textbf{38.78} & 0.992 & 0.045 & 0.388 & 38.72 & 0.992 & \textbf{0.041} & \textbf{0.288} \\
        [3pt]
        \textbf{~~Forward-facing scenes}  & & & & & & & &
         \\
        ~~~~~Hand & \textbf{26.84} & \textbf{0.834} & \textbf{0.246} & 0.388 & 26.72 & 0.832 & 0.253 & \textbf{0.104} \\
        ~~~~~Face & \textbf{33.91} & 0.951 & \textbf{0.215} & 0.384 & 33.13 & \textbf{0.953} & 0.222 & \textbf{0.102} \\
        [6pt]
        \textbf{~~Average} & \textbf{32.34} & \textbf{0.917} & \textbf{0.153} & 0.391 & 32.22 & 0.914 & 0.157 & \textbf{0.176} \\
        \hline
        \rule{0pt}{0ex} & & & & & & & & \\
        \textbf{~~Average (Overall)} & 27.18 & 0.8896 & 0.177 & 0.389 & \textbf{27.42} & \textbf{0.8899} & \textbf{0.175} & \textbf{0.176} \\
        \hline 
    \end{tabular}
        \caption{Quantitative evaluation of our \ac{tdg} method on datasets RGBT-Scenes and ThermalMix, compared to MSMG baseline.
    }
    \label{tab:datasets}
\end{table*}

\subsubsection{Progressive Joint Optimization}

Instead of maintaining separate models for different modalities, we optimize a single, unified 3D Gaussian model using a combined loss function. To address the training instability caused by the sparse texture and low contrast of thermal images, we propose a progressive joint optimization strategy. In the early stages of training, the 3D geometry is highly unconstrained. Therefore, we incorporate the depth rendering loss to enforce structural accuracy based on the estimated depth prior. As the training progresses and the overall geometry stabilizes, we gradually decay the weight of the depth loss, allowing the model to focus mainly on the photometric fidelity of the thermal rendering.

The progressive weight factor $w_{decay}(t)$ at iteration $t$ is formulated as a linear decay:
\begin{equation}
    \label{eq:progressive_weight}
    w_{decay}(t) = \max \left( 0, 1 - \frac{t - t_{start}}{t_{end} - t_{start}} \right),
\end{equation}
where $t_{start}$ and $t_{end}$ define the boundaries of the decay phase.

The final joint loss function is then formulated as follows:
\begin{equation}
    \mathcal{L}_{total} = \mathcal{L}_{\text{thermal}} + w_{decay}(t) \cdot \lambda_{depth} \cdot \mathcal{L}_{\text{depth}},
\end{equation}
where $\lambda_{depth}$ is a constant scaling hyperparameter for depth supervision. This progressive formulation ensures a smooth transition from geometry-guided structuring to fine-grained thermal optimization, effectively avoiding the optimization shocks associated with a hard supervision switch.

\section{Evaluation}

In the following, we first report our implementation details, followed by the evaluation strategy and our experiments.

\subsection{Implementation Details}

The code is written in Python using PyTorch, with CUDA/C++ components for efficient \ac{gs} rasterization. 
All experiments are conducted on a laptop equipped with an NVIDIA GeForce RTX 3060 GPU (6 GB VRAM), 16 GB RAM, and an Intel 12th Gen Core i7-12700H CPU (2.30 GHz). The implementation of the proposed framework, along with the experimental scripts and codes, are made available within the repository \textbf{\ac{tdg}}\footnote{Code available: \href{https://hannahhaensen.github.io/TDg/}{https://hannahhaensen.github.io/TDg/}}.

For data preprocessing, depth maps are estimated from thermal images using the method of Marigold~\citep{ke_repurposing_2024}. The images are posed and the initial sparse point clouds are obtained from COLMAP \citep{schonberger_structure--motion_2016} by using the \ac{sfm} algorithm. Since thermal images cannot provide the necessary geometric information to construct these point clouds, we leveraged cross-modal prior knowledge from RGB images. But our core \ac{tdg} method still uses only thermal images. Only for acquiring the sparse point clouds, we used RGB data.

We adopted the training settings from the official \ac{gs} implementation, keeping hyperparameters unchanged while adjusting the number of iterations to between 10k and 30k, as our method converges faster due to its simplified design. As a baseline, we select MSMG \citep{lu_thermalgaussian_2024}. For tuning the progressive weight factor~\eqref{eq:progressive_weight}, in our implementation, $t_{start}$ is the first iteration, and $t_{end}$ is typically set to 50\% of the total iterations to ensure that the depth prior is fully faded out in the later stages.

\subsection{Datasets}

We conducted comprehensive evaluations on the standard datasets used for thermal imaging and \ac{nvs}.

As datasets, we use RGBT-Scenes~\citep{lu_thermalgaussian_2024} and ThermalMix~\citep{ozer_exploring_2024}, following a standard protocol where 80\% of the images in each scene are used for training and 20\% for testing. We evaluated on 14 different scenes in total.

\paragraph{RGBT-Scenes:}
This dataset provides multi-view RGB images paired with pseudo-color thermal images, with a resolution of $640 \times 480$ ~\citep{lu_thermalgaussian_2024}. 
We evaluated on 9 scenes, including one low-light environment. 
The dataset is originally collected using a handheld thermal-infrared camera with accurate calibration, making it suitable for multimodal 3D reconstruction tasks.  
In addition, the pseudo-color thermal images can be further exploited for depth estimation during preprocessing.  

\paragraph{ThermalMix:}
This dataset contains paired RGB and grayscale thermal images of multiple indoor objects, also with a resolution of $640 \times 480$ ~\citep{ozer_exploring_2024}. 
We use 5 representative scenes from this dataset for evaluation. 
The precise RGB–Thermal alignment facilitates reliable 3D reconstruction, and the grayscale thermal images can also be directly employed for depth estimation in our pipeline.  

\subsection{Metrics}

For evaluation, rendering quality is measured using \ac{psnr}, \ac{ssim}, and \ac{lpips}, which together capture pixel-level fidelity, structural similarity, and perceptual consistency. Computational efficiency is reported as the average training time per scene in hours. Lower \ac{lpips} value, together with higher \ac{ssim} and \ac{psnr} values, signify better reconstruction fidelity. \ac{lpips} also indicates how visually appealing the results are to humans. \ac{ssim} and \ac{lpips} are especially important in low-contrast thermal scenarios.

\begin{figure}[t!]
\centering
\subfloat[Dark Scenes\label{fig:darkscenes}]{%
    \includegraphics[width=0.72\linewidth]{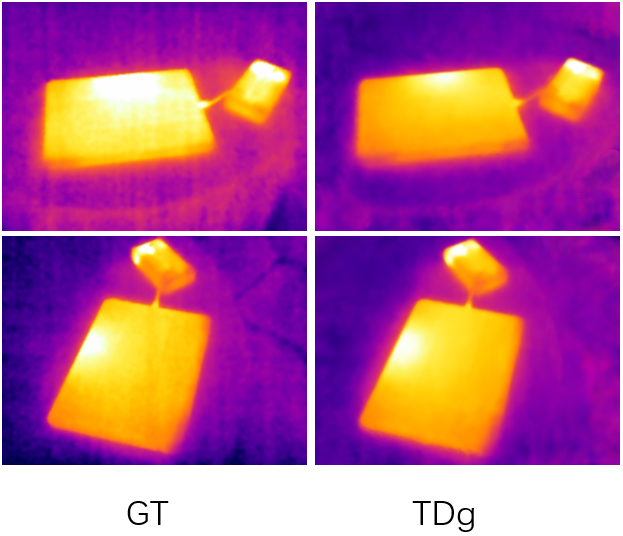}
    }
\vspace{0.8em}
\subfloat[Glass Cup\label{fig:glasscup}]{%
    \includegraphics[width=0.72\linewidth]{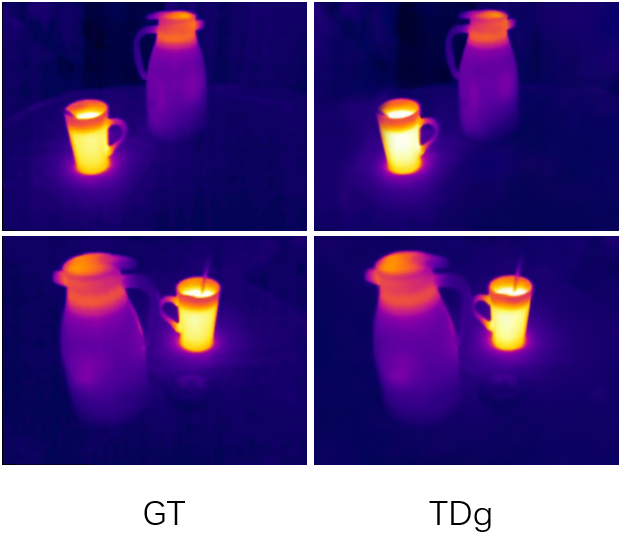}
    }
\caption{Rendering image results on the dataset scenes: (a) Dark Scenes (top) and (b) Glass Cup (bottom).}
\label{fig:rendering_results}
\end{figure}

\subsection{Quantitative and Qualitative Results}

We evaluate \ac{tdg} quantitatively and compare with state-of-the-art methods. Table~\ref{tab:datasets} shows a higher performance of \ac{tdg} compared to MSMG overall in mean \ac{psnr}, \ac{ssim}, \ac{lpips}, and runtime. For the runtime, overall we outperform the baseline, while for \ac{nvs} metrics we outperform MSMG on all scenes from the RGBT-Scenes dataset, except for one \ac{ssim} result. For the ThermalMix dataset, we also show an overall lower runtime across all scenes. However, for \ac{psnr}, \ac{ssim}, and \ac{lpips}, results are more mixed. Nevertheless, for \ac{psnr} we are within $\pm1.0$.

For qualitative results, we can visualize Figure~\ref{fig:rendering_results}, which shows rendering images on the datasets, Glass Cup and Dark Scenes. The thermal images rendered from our resulting GS model are compared against Ground Truth (GT), thereby showing the results of correct reconstruction. 

Overall, experimental results show that the depth-estimation-based approach improves reconstruction quality while maintaining accuracy, with a reduction in training time. It outperforms the baseline across various rendering quality metrics (\ac{psnr}, \ac{lpips}, and \ac{ssim}) as well as in terms of training efficiency. E.g., on average, the accuracy metric \ac{psnr} of our method is 27.42, which is better than the baseline value of 27.18. Another notable advantage of our \ac{tdg} method is the significant reduction in training time by 55\% (12 mins
47 secs), while still outperforming the baseline in \ac{nvs} quality. The runtime improvement is apparent in every scene, see Table \ref{tab:datasets}.

\subsection{Failure Cases}

\noindent\textbf{Random Point Cloud Initialization:} We experimented with initializing the 3D Gaussians from a random sparse point cloud, bypassing the \ac{sfm} requirement for point cloud generation. While this approach successfully minimized the photometric loss on the training set, it exhibited severe overfitting and failed to generalize to novel views. 

\begin{figure}[t!]
    \centering
    \includegraphics[width=0.97\linewidth]{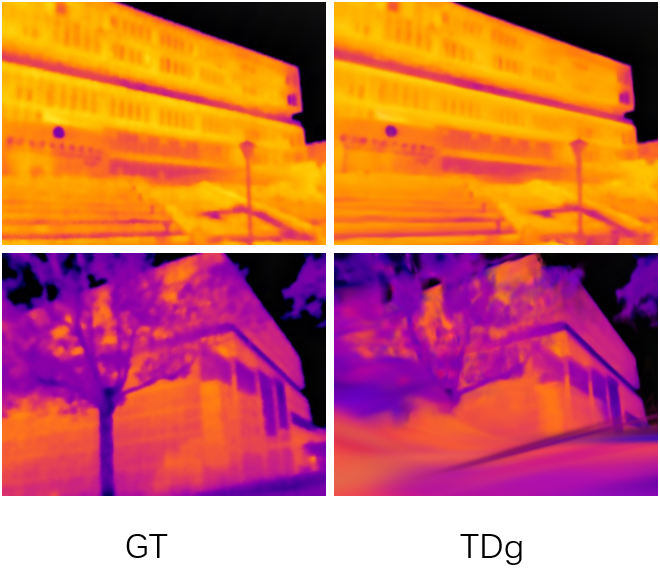}
\caption{%
Building reconstruction: The front view was correctly reconstructed, but the back view failed due to insufficient perspective.%
\label{fig:building}%
}
\end{figure}

\noindent\textbf{Partly Occluded Areas:} Additionally, our framework encounters difficulties when reconstructing large-scale targets, such as complete buildings. For such expansive structures, capturing a continuous and exhaustive 360-degree thermal multi-view trajectory is often logistically infeasible, leaving certain areas, such as the rear facades, sparsely observed. Figure~\ref{fig:building} shows such an example of a building, where the front view was correctly reconstructed, but the back view failed due to insufficient perspective. While our depth-guided \ac{tdg} approach effectively regularizes the geometry in observed regions, it cannot properly densify Gaussians in rear-seen spaces. 

\section{Discussion}

In summary, we contribute a thermal-only \ac{gs} approach, outperforming the baseline in training time and \ac{nvs} quality.

A fundamental challenge in jointly optimizing thermal and RGB modalities is the inherent discrepancy in their physical properties. RGB data is highly sensitive to ambient lighting variations and high-frequency texture noise, which can introduce conflicting gradient signals during the optimization of thermal Gaussians. During the training of thermal Gaussians, it can sometimes confuse the model rather than helping it. Furthermore, forcing the model to accommodate redundant, low-quality, or misaligned RGB features can distract the optimization process, ultimately degrading the representation of critical thermal signatures. It can reduce the focus on key thermal features. This justifies our thermal-centric pipeline, which avoids these cross-modal conflicts during the core rendering phase.

For runtime efficiency, we outperform the multi-modal baseline due to an efficiency gain. This efficiency gain is directly attributed to the depth-guided supervision. In the baseline MSMG, the lack of geometric constraints causes the 3D Gaussians to perform excessive and inefficient densification in empty spaces, to minimize the photometric loss, resulting in \enquote*{floaters}. By incorporating the depth rendering loss ($\mathcal{L}_{depth}$), \ac{tdg} provides an explicit spatial gradient that quickly guides the Gaussians to the correct geometry planes, drastically reducing unnecessary splitting and pruning operations, thereby accelerating the overall convergence.

\subsection{Limitations}

Currently, our model still requires a sparse point cloud for initialization. Our initialization step relies on an RGB-based \ac{sfm} producing a COLMAP point cloud. Because the low resolution of thermal images poses challenges for the \ac{sfm} algorithm to carry out tasks such as feature extraction, matching descriptors, etc.

However, \ac{gs} also allows random sparse point cloud initialization, instead of using points already in a proper geometric format. While random point clouds can be used, the \ac{nvs} quality degrades. Notably, also for the baseline. This failure case confirms that without reliable geometric priors, purely thermal-based optimization easily falls into local minima, thereby justifying our design choice of utilizing RGB-assisted \ac{sfm} for the initial geometric scaffolding. 

As shown in Table \ref{tab:point_cloud}, our approach, as well as the baseline, relies on a proper geometric initialization. An interesting observation was that jointly training thermal and RGB splats with random point cloud initialization also deteriorates the quality of rendered RGB images.

\subsection{Future Work}

Future work can focus on completely removing the reliance on RGB data, by exploring the possibility of using random sparse point clouds as initialization. This will also reduce the reliance on algorithms such as \ac{sfm}. Or alternatively, perform temporally consistent normalization and local detail enhancement on thermal images to get better features for point cloud initialization. In this way, we can move towards adapting existing feature extraction and matching approaches for RGB images to thermal images and do an end-to-end reconstruction, using only thermal images. And it will also remove the constraint of performing RGB-thermal registration or alignment. Another scope of future study is to further improve the reliability and accuracy of depth estimation from thermal images.

\begin{table}[t!]
    \centering 
    \large
    \renewcommand{\arraystretch}{1.387}
    \resizebox{\columnwidth}{!}{
    \begin{tabular}{|l|cc|c|} \hline 
         & Random & COLMAP & \ac{psnr} / \ac{ssim} \\ \hline
       MSMG~\citep{lu_thermalgaussian_2024} & $\checkmark$ & $\times$ & 9.64 / 0.3471\\
       MSMG~\citep{lu_thermalgaussian_2024}  & $\times$ & \checkmark & \textbf{27.18 / 0.8896}\\ \hline
       Ours (\ac{tdg}) & \checkmark & $\times$ &   9.69 / 0.3472\\
       Ours (\ac{tdg}) & $\times$ & \checkmark & \textbf{27.42 / 0.8899}  \\ \hline 
    \end{tabular}}
    \caption{Ablation study on RGBT-Scenes and ThermalMix dataset, we report mean overall \ac{psnr} and \ac{ssim}. We compare random initialization and COLMAP-based initialization of the point cloud used to start the \ac{gs} training.}
    \label{tab:point_cloud}
\end{table}

\section{Conclusion}
In this paper, we have presented a thermal-guided depth estimation module, \acf{tdg}, that uses thermal images and depth estimation in its architecture to derive the radiance fields, to accurately represent scenes. Experimental results show that our method performs better than the MSMG baseline across various rendering quality metrics and in terms of training efficiency, when tested on standard datasets. The accuracy metric \ac{psnr} of our method is 0.01\% better than the baseline value, and the training time reduces by 12 mins 47 secs (55\% reduction). From the training part, we can summarize that thermal images alone are powerful. Even without RGB, they provide strong structural and semantic signals. Estimated depth can be useful. Despite its noise, it helps when guided by well-crafted loss terms.

Therefore, overall, our method is successful in deriving the 3D thermal models in the form of radiance fields, which can ultimately have several applications, such as identifying heat sources (e.g., people, machinery) critical in surveillance, search and rescue operations, or industrial inspection. They can be suitable for robust robot vision regardless of lighting and weather conditions, and can be used in factories where temperature is widely used to monitor machines. Furthermore, an additional advantage is that there is no need for multi-modal data and hardware synchronization.

\bibliography{ISPRSguidelines_authors} 

\end{document}

%% file: acronym.tex
\begin{acronym}[Bspwwww.]  

\acro{ar}[AR]{augmented reality}
\acro{ap}[AP]{average precision}
\acro{api}[API]{application programming interface}
\acroplural{ann}[ANN]{artifical neural networks}

\acro{tdg}[TDg]{Thermal-to-Depth Gaussian Splatting}

\acro{bev}[BEV]{bird eye view}
\acro{rbob}[BRB]{Bottleneck residual block}
\acroplural{rbob}[BRBs]{Bottleneck residual blocks}
\acro{mbiou}[mBIoU]{mean Boundary Intersection over Union}
\acro{poi}[POI]{Point of Interest}
\acro{cai}[CAI]{computer-assisted intervention}
\acro{ce}[CE]{cross entropy}
\acro{cad}[CAD]{computer-aided design}
\acro{cnn}[CNN]{convolutional neural network}

\acro{crf}[CRF]{conditional random fields}
\acro{dpc}[DPC]{dense prediction cells}
\acro{dla}[DLA]{deep layer aggregation}
\acro{dnn}[DNN]{deep neural network}
\acroplural{dnn}[DNNs]{deep neural networks}

\acro{da}[DA]{domain adaption}
\acro{dr}[DR]{domain randomization}
\acro{fat}[FAT]{falling things}
\acro{fcn}[FCN]{fully convolutional network}
\acroplural{fcn}[FCNs]{fully convolutional networks}
\acro{fov}[FoV]{field of view}
\acro{fv}[FV]{front view}
\acro{fp}[FP]{False Positive}
\acro{fpn}[FPN]{feature Pyramid network}
\acro{fn}[FN]{False Negative}
\acro{fmss}[FMSS]{fast motion sickness scale}
\acro{gan}[GAN]{generative adversarial network}
\acroplural{gan}[GANs]{generative adversarial networks}
\acro{gcn}[GCN]{graph convolutional network}
\acroplural{gcn}[GCNs]{graph convolutional networks}
\acro{gs}[GS]{Gaussian Splatting}
\acro{gg}[GG]{Gaussian Grouping}
\acro{hmi}[HMI]{Human-Machine-Interaction}
\acro{hmd}[HMD]{Head Mounted Display}
\acroplural{hmd}[HMDs]{head mounted displays}
\acro{iou}[IoU]{intersection over union}
\acro{irb}[IRB]{inverted residual bock}
\acroplural{irb}[IRBs]{inverted residual blocks}
\acro{ipq}[IPQ]{igroup presence questionnaire}
\acro{knn}[KNN]{k-nearest-neighbor}
\acro{lidar}[LiDAR]{light detection and ranging}
\acro{lsfe}[LSFE]{large scale feature extractor}
\acro{llm}[LLM]{large language model}
\acro{map}[mAP]{mean average precision}
\acro{mc}[MC]{mismatch correction module}
\acro{miou}[mIoU]{mean intersection over union}
\acro{mis}[MIS]{Minimally Invasive Surgery}
\acro{msdl}[MSDL]{Multi-Scale Dice Loss}
\acro{ml}[ML]{Machine Learning}
\acro{mlp}[MLP]{multilayer perception}
\acro{miou}[mIoU]{mean Intersection over Union}
\acro{nn}[NN]{neural network}
\acroplural{nn}[NNs]{neural networks}
\acro{ndd}[NDDS]{NVIDIA Deep Learning Data Synthesizer}
\acro{nocs}[NOCS]{Normalized Object Coordiante Space}
\acro{nerf}[NeRF]{Neural Radiance Fields}
\acro{NVISII}[NVISII]{NVIDIA Scene Imaging Interface}
\acro{ngp}[NGP]{Neural Graphics Primitives}
\acro{or}[OR]{Operating Room}
\acro{pbr}[PBR]{physically based rendering}
\acro{psnr}[PSNR]{peak signal-to-noise ratio}
\acro{pnp}[PnP]{Perspective-n-Point}
\acro{rv}[RV]{range view}
\acro{roi}[RoI]{region of interest}
\acroplural{roi}[RoIs]{region of interests}
\acro{rbab}[BB]{residual basic block}
\acro{ras}[RAS]{robot-assisted surgery}
\acroplural{rbab}[BBs]{residual basic blocks}
\acro{spp}[SPP]{spatial pyramid pooling}
\acro{sh}[SH]{spherical harmonics}
\acro{sgd}[SGD]{stochastic gradient descent}
\acro{sdf}[SDF]{signed distance field}
\acro{sfm}[SfM]{Structure-from-Motion}
\acro{sam}[SAM]{Segment-Anything}
\acro{sus}[SUS]{system usability scale}
\acro{ssim}[SSIM]{structural similarity index measure}

\acro{slam}[SLAM]{simultaneous localization and mapping}
\acro{tp}[TP]{True Positive}
\acro{tn}[TN]{True Negative}
\acro{thor}[thor]{The House Of inteRactions}
\acro{tsdf}[TSDF]{truncated signed distance function}
\acro{vr}[VR]{Virtual Reality}
\acro{ycb}[YCB]{Yale-CMU-Berkeley}

\acro{ar}[AR]{augmented reality}
\acro{ate}[ATE]{absolute trajectory error}
\acro{bvip}[BVIP]{blind or visually impaired people}
\acro{cnn}[CNN]{convolutional neural network}
\acro{c2f}[c2f]{coarse-to-fine}
\acro{fov}[FoV]{field of view}
\acro{gan}[GAN]{generative adversarial network}
\acro{gcn}[GCN]{graph convolutional Network}
\acro{gnn}[GNN]{Graph Neural Network}
\acro{hmi}[HMI]{Human-Machine-Interaction}
\acro{hmd}[HMD]{head-mounted display}
\acro{mr}[MR]{mixed reality}
\acro{iot}[IoT]{internet of things}
\acro{llff}[LLFF]{Local Light Field Fusion}
\acro{bleff}[BLEFF]{Blender Forward Facing}

\acro{lpips}[LPIPS]{learned perceptual image patch similarity}
\acro{nerf}[NeRF]{neural radiance fields}
\acro{nvs}[NVS]{novel view synthesis}
\acro{mlp}[MLP]{multilayer perceptron}
\acro{mrs}[MRS]{Mixed Region Sampling}

\acro{or}[OR]{Operating Room}
\acro{pbr}[PBR]{physically based rendering}
\acro{psnr}[PSNR]{peak signal-to-noise ratio}
\acro{pnp}[PnP]{Perspective-n-Point}
%
\acro{sus}[SUS]{system usability scale}

\acro{slam}[SLAM]{simultaneous localization and mapping}

\acro{tp}[TP]{True Positive}
\acro{tn}[TN]{True Negative}
\acro{thor}[thor]{The House Of inteRactions}
\acro{ueq}[UEQ]{User Experience Questionnaire}
\acro{vr}[VR]{virtual reality}
\acro{who}[WHO]{World Health Organization}
\acro{xr}[XR]{extended reality}
\acro{ycb}[YCB]{Yale-CMU-Berkeley}
\acro{yolo}[YOLO]{you only look once}
\end{acronym}